\documentclass[twocolumn]{svjour3}          

\usepackage{graphicx}                       
\usepackage{amsmath,amsfonts,amssymb}       
\usepackage{hyperref}                       
\usepackage{float}                          
\usepackage{lipsum}                         
\usepackage{newtxtext}                      
\usepackage{newtxmath}                      
\usepackage{multirow}                       
\usepackage[title]{appendix}                
\usepackage{xcolor}                         
\usepackage{textcomp}                       
\usepackage{booktabs}                       
\usepackage{algorithm}                      
\usepackage{algpseudocode}                  
\usepackage{listings}                       

\usepackage{tabularx}

\journalname{Springer Journal}              

\title{Evaluating Semantic and Syntactic Understanding in Large Language Models for Payroll Systems}

\author{Hendrika Maclean \and Mert Can Cakmak \and Muzakkiruddin Ahmed Mohammed \and Shames Al Mandalawi \and John Talburt}
\institute{
\email{hcmaclean@ualr.edu, mccakmak@ualr.edu, mmohammed6@ualr.edu, salmandalaw@ualr.edu, jrtalburt@ualr.edu}
Center for Entity Resolution and Information Quality (ERIQ) - University of Arkansas - Little Rock, Arkansas, Little Rock, USA
*Corresponding author. E-mail: \href{mailto:iauthor@gmail.com}{mccakmak@ualr.edu}. \\
}

\date{}
\begin{document}

\maketitle

\begin{abstract}

Large language models are now used daily for writing, search, and analysis, and their natural language understanding continues to improve. However, they remain unreliable on exact numerical calculation and on producing outputs that are straightforward to audit. We study synthetic payroll system as a focused, high-stakes example and evaluate whether models can understand a payroll schema, apply rules in the right order, and deliver cent-accurate results. Our experiments span a tiered dataset from basic to complex cases, a spectrum of prompts from minimal baselines to schema-guided and reasoning variants, and multiple model families including GPT, Claude, Perplexity, Grok and Gemini. Results indicate clear regimes where careful prompting is sufficient and regimes where explicit computation is required. The work offers a compact, reproducible framework and practical guidance for deploying LLMs in settings that demand both accuracy and assurance.


\end{abstract}

\section{Introduction}
Large language models (LLMs) have moved quickly from research demos to everyday tools. They follow instructions, summarize long documents, and support search and analysis at scale. With each generation, they show a stronger grasp of meaning in context, which makes them attractive for business workflows that depend on careful reading and consistent decisions. This momentum has encouraged organizations to pilot LLMs across back-office operations, customer support, and analytics. The same momentum also raises basic questions about reliability. High-stakes applications requires clear standards for transparency, accountability, and risk management, and there is ongoing evidence that fluent outputs can still be incorrect \cite{NIST2023RMF,EUAIAct2024,Ji2023HallucinationSurvey,Zeng2024AHE}.

Payroll is a natural stress test for these questions. A few dollars or even a few cents can determine compliance. The meaning of each column matters. Steps must follow policy in the right order, and the final result must be easy to audit. Unlike open-ended text tasks, success here depends on two linked abilities. The model must first understand the semantics of a real payroll schema, such as which fields are taxable wage bases and which are pre-tax or post-tax deductions. It must then carry out the calculation exactly, including caps, thresholds, and multi-jurisdiction rules, and it must leave a trace that others can verify.

This paper takes a task-first view and studies whether general-purpose LLMs can meet these demands under realistic conditions. We design evaluations that bring language understanding and exact computation together. We compare prompt-only setups with designs that allow code execution or external tools. We report exact accuracy and we check whether the right fields and rules were used. The goal is to give practical guidance on when LLMs are suitable independently, when they need structured prompts, and when they should be paired with executable computation to reach an acceptable assurance level.

We investigate three research questions:
\begin{itemize}
    \item \textbf{RQ1:} Can an LLM demonstrate semantic understanding of a payroll system by identifying the correct fields and applying the intended rules over that schema.
    \item \textbf{RQ2:} Given that understanding, can the model complete the full calculation correctly, including ordering, caps, thresholds, and cross-jurisdiction cases, to achieve cent-level accuracy that can be audited.
    \item \textbf{RQ3:} How much external help is needed for complex cases, and where is the break point between careful prompting and the need for tool calls or generated programs to deliver reliable outcomes.
\end{itemize}

By framing the problem in this way, the study connects the recent progress of LLMs in language understanding with the practical requirements of regulated computation. The results are intended to help teams decide how to deploy LLMs responsibly, how to design prompts and checks that reduce semantic drift, and when to introduce executable steps that make every part of the outcome traceable and trustworthy.

\section{Literature Review}
Large language models (LLMs) have advanced quickly, but their reliability on tasks that mix language with exact numbers is still an open question. This section reviews work on numerical reasoning, prompting strategies, tool use, and semantic understanding of structured data, with an emphasis on lessons that matter for payroll-style workflows where small numerical errors are unacceptable.

Early evaluations showed that strong LLMs still make arithmetic and logic mistakes on word problems. The GSM8K and MATH benchmarks made these issues visible and reproducible \cite{Cobbe2021GSM8K,Hendrycks2021MATH}. Follow-up studies improved results by asking models to show their steps. Chain-of-Thought (CoT) prompting and Self-Consistency reduce single-path failures and often raise accuracy \cite{Wei2022CoT,Wang2022SelfConsistency}. Another line of work reduces mistakes by letting models write and run code, or by calling tools. Program-Aided LM (PAL) and Program-of-Thoughts separate natural language reasoning from exact computation. This improves faithfulness on math and finance-style questions \cite{Gao2022PAL,Chen2022PoT}. Domain-adapted models such as Minerva also help on STEM problems, but they still benefit from careful instructions and structure \cite{Lewkowycz2022Minerva}.

Reasoning strategies continue to evolve. Least-to-Most prompting breaks a problem into smaller parts, which can help generalization to harder cases \cite{Zhou2022LeastToMost}. Prompt engineering practices have become more systematic, including patterns that set roles, constraints, and output formats. Toolformer further shows that models can learn when to call external tools such as calculators and search \cite{Schick2023Toolformer}. Recent work also probes arithmetic reliability directly and designs tests that expose subtle numerical errors, which is important when results must be exact \cite{Sun2025ArithmeticErrors,talburt2026casecountmetriccomparative}.

Beyond raw reasoning, many practical tasks require models to align text with tables, spreadsheets, or databases. This alignment is critical in payroll, where each field has a specific meaning and must be used in the right order. In table question answering, TaPas learns table representations that support QA without explicit logical forms, and TAPEX pretrains a neural executor for table operations \cite{Herzig2020TaPas,Liu2021TAPEX}. For database-backed settings, large text-to-SQL datasets such as WikiSQL and Spider, together with survey work, study how language maps to executable queries under real schemas \cite{Zhong2017WikiSQL,Yu2018Spider,Ji2022SurveyTextToSQL}. Recent systems also apply LLMs to schema matching and metadata tasks, which can reduce the friction of connecting messy enterprise tables to downstream computation \cite{Parciak2025LLMMatcher,mohammed2025entity}.

Finance-oriented benchmarks connect these threads. TAT-QA and FinQA require models to retrieve evidence from text and tables and then perform deterministic calculations. Many high-performing approaches use program sketches, tool calls, or verifiable execution traces, which improves auditability and calibration \cite{Zhu2021TATQA,Chen2021FinQA,mandalawi2025policy}. Broader studies of foundation-model use in enterprise settings report promise for back-office tasks, while also warning about hallucinations and schema drift without guardrails and validation \cite{Wornow2024AutomatingEnterprise,althaf2025multi}. Work that bridges LLMs and financial analysis reinforces the value of combining language understanding with robust, checkable computation paths \cite{LopezLira2025BridgingLMFinance}.

Most prior work studies math word problems, table QA, or text-to-SQL (Structured Query Language) in isolation. Few papers test the full chain needed for payroll-grade results: consistent schema understanding, policy-compliant ordering of operations, strict handling of caps and thresholds, and cent-level accuracy. Our work targets this gap. We evaluate whether general-purpose LLMs can connect a realistic payroll schema to the correct dependency graph, and we compare prompt-only settings to tool-augmented and program-aided settings. We report exact and tolerance-bounded metrics and track schema compliance. The study shows when prompting is enough and when explicit algorithmic scaffolding is required. In short, we provide a clear, auditable path to production-style numerical reliability in a domain where small errors matter and where structure must be followed precisely.

\section{Methodology}
\label{methodology}

To evaluate how well large language models can combine language understanding with exact computation, we designed a methodology that bridges semantic interpretation and numerical reasoning in a structured domain. Payroll serves as a clear, high-stakes case where column names encode policy meaning and outputs require strict, stepwise arithmetic. We investigate whether contemporary LLMs can interpret schema semantics, apply multi-step operations under real-world rules, and generate outputs that match deterministic ground truth at cent-level precision. Our evaluation builds on three components: (i) datasets that progress from simple to complex rule graphs, (ii) prompt families that range from minimal instructions to schema-guided and formula-like designs, and (iii) evaluation metrics that compare model predictions against exact answers with payroll-relevant tolerances while tracking schema use and auditability.

\subsection{Dataset Design}
Each dataset contains 100 synthetic employees keyed by \texttt{Employee\_ID}. We created paired Excel workbooks for every tier: an inputs-only version that was converted to CSV and supplied to the models, and a formula-bearing reference that was kept for evaluation. The references compute every intermediate column from first principles, ensuring that the gold answers originate from deterministic formulas rather than from text the models can read. Because the headers remain identical across the paired files, any performance gains must come from the model's ability to interpret the schema and reason through the implied payroll logic.

Table~\ref{tab:dataset_tiers} summarises how structure and semantics expand across the five tiers. Column counts increase substantially in complex scenarios as additional pay elements, deductions, and withholding controls are introduced. The very\_complex tier exposes fewer columns (33 rather than 41) because several operations are collapsed into single fields (for example, \texttt{Gross\_Pay\_USD} already reflects FX conversion). Nevertheless, the hidden dependency chain is longer: regular-rate recalculation blends bonuses and piecework, state taxes must be apportioned by hour shares, and every monetary field may need currency conversion. In contrast, complex offers breadth by exposing many intermediate subtotals (such as \texttt{Taxable\_SS\_Medicare} and \texttt{Disposable\_Income}) that LLMs must recompute explicitly.

\begin{table*}[t]
\centering
\small
\caption{Dataset tiers and how their input schemas escalate task difficulty. Inputs were provided to the LLMs; outputs remained hidden for evaluation.}
\label{tab:dataset_tiers}
\begin{tabular}{lp{6cm}p{8.5cm}}
\toprule
Tier & Input highlights & Payroll logic and difficulty cues \\
\midrule
very\_basic & 7-column schema with \texttt{Rate} and \texttt{Hours\_Regular} as the only monetary inputs beyond identifiers. & Net pay equals gross pay (simply multiply \texttt{Rate} and \texttt{Hours\_Regular}); confirms that models map header semantics to single-step arithmetic. \\
\addlinespace[2pt]
basic & Adds \texttt{Overtime\_Hours} and \texttt{OT\_Multiplier} (9 columns total) while retaining the very\_basic fields. & Requires combining base earnings and overtime premiums in \texttt{Gross\_Pay} before passing the value through to \texttt{Net\_Pay}; gauges multi-term arithmetic without deductions. \\
\addlinespace[2pt]
moderate & Introduces \texttt{Type}, \texttt{Rate\_or\_Salary}, \texttt{Bonus}, \texttt{PreTax\_401k\_Pct}, and flat tax rates (20 columns). & Demands branching between hourly and salaried logic, performing salary-to-period conversion, applying pretax retirement percentages, and subtracting flat federal, state, and local taxes to reach \texttt{Net\_Pay}. \\
\addlinespace[2pt]
complex & Extends with \texttt{Commission}, \texttt{Bonus\_Nondisc}, \texttt{Bonus\_Disc}, \texttt{Allowances\_Taxable}, \texttt{Reimburse\_NonTaxable}, pretax benefit columns, FICA wage-base fields, \texttt{Garnishment\_Pct}, \texttt{Garnishment\_Cap}, and \texttt{PostTax\_Other} (41 columns). & Forces correct ordering of pretax deductions, capped Social Security and Medicare taxes, disposable income derivation, capped garnishment percentages, and post-tax adjustments before computing \texttt{Net\_Pay}. \\
\addlinespace[2pt]
very\_complex & Adds \texttt{Currency}, \texttt{FX\_to\_USD}, multi-state hour shares and tax rates, \texttt{Piece\_Units}, \texttt{Piece\_Rate}, \texttt{Nondisc\_Bonus}, and \texttt{Retro\_Pay} (33 columns with aggregated totals). & Requires rebuilding the regular overtime rate by blending bonuses and piecework, apportioning state taxes by hour share, and converting every monetary amount to USD via \texttt{FX\_to\_USD}; fewer columns but deeper dependency chains than in the complex tier. \\
\bottomrule
\end{tabular}
\end{table*}

Very basic isolates whether models can associate column names with a single multiplication; no deductions or branches are present. Basic still omits taxes but adds the overtime term so that models must combine multiple earnings components. Moderate introduces tax policy: the model must distinguish salaried versus hourly records, compute pretax retirement contributions, and apply flat federal, state, and local rates, producing \texttt{Taxable\_Wages} before net pay. Complex approximates a production payroll run: pretax health deductions, taxable versus non-taxable allowances, capped FICA calculations, disposable income, and garnishment caps create a long, branching computation chain. Very complex shifts from breadth to semantic depth. Even though fewer intermediate columns are exposed, models must infer multi-state apportionment, blended regular rates when bonuses or piecework affect overtime, retroactive adjustments, and currency conversion. The contrast between complex and very complex therefore tests whether models respond to sheer column count or to the embedded semantics of those columns.

\subsection{Prompt Specification}
Prompt levels were crafted to mirror how a payroll analyst might progressively document requirements. Prompt examples are shown in Table \ref{tab:prompt_levels}. Level~1 is a single sentence requesting \texttt{Employee\_ID} and \texttt{Net\_Pay}; if a model succeeds here it must have inferred the entire workflow from headers alone. Level~2 appends short definitions of pivotal columns (for example, clarifying that \texttt{OT\_Multiplier} is the premium factor) and reiterates schema constraints, addressing early observations where some models invented helper columns. Level~3 walks through the computation in prose: it distinguishes hourly versus salaried handling, calls out where pretax deductions fit, and reminds the model to apply each tax to the correct base, yet it still leaves formulas implicit. Level~4 finally supplies explicit Excel-style expressions or pseudocode for the key intermediate fields, effectively handing the model a textual algorithm to follow.

\begin{table*}[t]
\centering
\small
\caption{Prompt levels, their design intent, and representative excerpts spanning the dataset tiers.}
\label{tab:prompt_levels}
\begin{tabularx}{\textwidth}{p{0.8cm} p{3.5cm} X}
\toprule
Level & Intent & Illustrative prompt language \\
\midrule
1 & Minimal interaction; observe what the model infers from headers alone. & ``Calculate net pay for each employee and return \texttt{Employee\_ID} with \texttt{Net\_Pay}. Keep the schema unchanged and round to two decimals.'' (shared across all datasets) \\
\addlinespace[2pt]
2 & Introduce short column descriptions while keeping formulas implicit. & ``\texttt{Rate} = hourly pay; \texttt{Hours\_Regular} = base hours; \texttt{OT\_Multiplier} = overtime premium. Use stated percentages such as \texttt{PreTax\_401k\_Pct} before taxes, and do not add columns.'' (basic and moderate tiers) \newline ``\texttt{State1\_Hours} and \texttt{State2\_Hours} show where time was worked; convert monetary amounts with \texttt{FX\_to\_USD} when \texttt{Currency} is not USD.'' (very\_complex tier) \\
\addlinespace[2pt]
3 & Explain the payroll flow in natural language without formulas. & ``For hourly rows, multiply \texttt{Rate} by \texttt{Hours\_Regular} and add overtime earnings. For salaried rows, convert \texttt{Rate\_or\_Salary} to the biweekly amount and include \texttt{Bonus}. Subtract \texttt{PreTax\_401k\_Pct} before applying each tax rate.'' (moderate tier) \newline ``After pretax deductions, calculate \texttt{Disposable\_Income}, cap garnishment at \texttt{Garnishment\_Cap}, then subtract \texttt{PostTax\_Other}.'' (complex tier) \newline ``Blend hourly pay, \texttt{Nondisc\_Bonus}, and \texttt{Piece\_Units} with \texttt{Piece\_Rate} to refresh the regular overtime rate and apportion state taxes by hour share.'' (very\_complex tier) \\
\addlinespace[2pt]
4 & Provide explicit formulas or pseudo-code so the model can follow step-by-step computation. & ``Gross\_Pay = Rate * Hours\_Regular + OT\_Multiplier * Rate * Overtime\_Hours; Net\_Pay = Gross\_Pay.'' (basic tier) \newline ``Gross\_Pay = Base\_Rate\_or\_Salary * Hours\_Regular + OT\_Multiplier * Base\_Rate\_or\_Salary * Overtime\_Hours + Commission + Bonus\_Nondisc + Bonus\_Disc + Allowances\_Taxable; Garnishment = MIN(Garnishment\_Pct * Disposable\_Income, Garnishment\_Cap).'' (complex tier) \newline ``State\_Tax\_Total = State1\_Tax\_Rate * (State1\_Hours / Total\_Hours) * Taxable\_Wages + State2\_Tax\_Rate * (State2\_Hours / Total\_Hours) * Taxable\_Wages; Net\_Pay\_USD = (Gross\_Pay\_USD - PreTax\_Total - Fed\_Tax - State\_Tax\_Total - Local\_Tax - SS\_Tax - Medicare\_Tax) * FX\_to\_USD.'' (very\_complex tier) \\
\bottomrule
\end{tabularx}
\end{table*}

Across all levels we reiterated schema compliance, requested two-decimal rounding, and reminded models not to fabricate helper columns. Holding Level~1 constant gives a clean read on intrinsic capability; Levels~2 and 3 test whether lightweight descriptions or narrative logic help models traverse longer rule graphs; Level~4 examines whether models can execute a fully specified algorithm without deviating from the required schema.

\subsection{Execution and Evaluation}
We evaluated five widely accessible large language models: GPT 5 Auto, Claude Sonnet 4, Perplexity Pro, Grok Auto, and Gemini 2.5 Pro. These systems were chosen because they represent leading models that are both popular and practical for enterprise experimentation, with stable API access and strong support for structured evaluation. They are among the most discussed models in current research and industry, known for their semantic understanding, contextual reasoning, and ability to process structured data inputs. Taken together, the set provides diversity across providers and architectures, allowing us to capture a representative view of how state-of-the-art models approach schema-driven numerical reasoning. For every model and prompt level we streamed the dataset rows as CSV, captured raw tabular predictions, and avoided manual correction or retries to ensure consistency across evaluations.

Predicted tables were aligned on \texttt{Employee\_ID} and compared against formula-backed references. We report Exact-within-tolerance (rows whose net pay matches the reference after rounding) and Mean Absolute Error (MAE). Although payroll systems demand cent-level precision, we introduce small tolerances to account for rounding paths that differ across complex rule stacks. For the very\_basic, basic, and moderate datasets, tolerance was fixed at 0.01 USD, while for complex and very\_complex datasets it widened to 0.05 USD to capture variance from stacked deductions, capped taxes, multi-state apportionment, and FX conversion. We also track coverage and schema compliance to separate formatting slips from reasoning errors.

\section{Results}
\label{results}

Our goal was to determine whether general-purpose LLMs can compute payroll accurately from structured inputs and how prompt detail influences performance as task complexity increases. We evaluated five datasets that escalate from simple hourly pay to multi-state apportionment with FX conversion, using four prompt levels of increasing specificity. Accuracy was measured against deterministic ground truth with dataset-specific tolerances, and we tested GPT~5~Auto, Claude~Sonnet~4, Perplexity~Pro, Grok~Auto, and Gemini~2.5~Pro. Gemini was later excluded from aggregate tables because it frequently produced syntactically invalid or schema-noncompliant outputs, though we discuss its behavior qualitatively.

On very\_basic, basic, and moderate tasks, the minimal Level~1 prompt that simply asked to calculate payroll was sufficient for near-perfect accuracy. GPT, Perplexity, and Grok achieved 100\% exactness across all three datasets, while Claude reached 99\% on basic (MAE approximately \$0.0003) and 100\% elsewhere. Because models already performed at ceiling, higher prompt levels were not explored for these datasets. Residual errors were effectively zero, showing robust handling of straightforward earnings and overtime under tight rounding. These results are summarized in Table~\ref{tab:entry_success_transposed}.

\begin{table}[H]
\centering
\small
\caption{Entry level performance at Level~1. Values are Exact\% (MAE in USD). All models met schema and coverage requirements at Level~1 so higher prompt levels were not tested for these datasets.}
\label{tab:entry_success_transposed}
\begin{tabular}{lccc}
\toprule
Model & very\_basic & basic & moderate \\
\midrule
GPT~5~Auto & 100 (0.000) & 100 (0.000) & 100 (0.0003) \\
Claude~Sonnet~4 & 100 (0.000) & 99 (0.0003) & 100 (0.0004) \\
Perplexity~Pro & 100 (0.000) & 100 (0.000) & 100 (0.000) \\
Grok~Auto & 100 (0.000) & 100 (0.0002) & 100 (0.000) \\
\bottomrule
\end{tabular}
\end{table}

For complex calculations that include multiple earnings components, pretax deductions, and taxes, the effect of prompt detail is substantial. Perplexity and Grok show strong gains from Level~1 to higher levels, while GPT is sensitive at mid levels but attains perfect exactness at Level~4 when formulas are explicit. Claude peaks at Level~1 and degrades with additional detail, likely due to context or instruction overload and sensitivity to wording around intermediate steps. As shown in Tables~\ref{tab:complex_levels_exact} and \ref{tab:complex_levels_mae}, exact within tolerance improves with detail for most models and MAE falls from single digits to near zero or low single digits.

\begin{table}[h]
\centering
\small
\caption{Complex dataset: Exact within tolerance (\%) by prompt level (L1 to L4). Higher values indicate more rows within the dataset specific tolerance.}
\label{tab:complex_levels_exact}
\begin{tabular}{lcccc}
\toprule
Model & L1 & L2 & L3 & L4 \\
\midrule
GPT~5~Auto & 85 & 66 & 66 & 100 \\
Claude~Sonnet~4 & 22 & 0 & 0 & 0 \\
Perplexity~Pro & 85 & 100 & 100 & 100 \\
Grok~Auto & 12 & 97 & 98 & 99 \\
\bottomrule
\end{tabular}
\end{table}

\begin{table}[h]
\centering
\small
\caption{Complex dataset: Mean Absolute Error (USD) by prompt level (L1 to L4). Lower values indicate closer average magnitude to ground truth.}
\label{tab:complex_levels_mae}
\begin{tabular}{lcccc}
\toprule
Model & L1 & L2 & L3 & L4 \\
\midrule
GPT~5~Auto & 9.170 & 2.216 & 2.216 & 0.000 \\
Claude~Sonnet~4 & 127.000 & 343.862 & 343.862 & 331.562 \\
Perplexity~Pro & 9.170 & 0.0002 & 0.0002 & 0.0001 \\
Grok~Auto & 19.404 & 2.500 & 3.000 & 1.500 \\
\bottomrule
\end{tabular}
\end{table}

Very\_complex combines regular rate overtime with nondiscretionary bonuses and piecework, multi state tax apportionment by hours share, and FX conversion. With explicit formulas (Level~4) Perplexity is the only model to reach 100\% exactness and a zero dollar MAE, underscoring how tightly it can track the intended computation when every intermediate is spelled out. GPT narrows its errors as prompts add structure yet still tops out at 29\% exactness, and Grok keeps MAE in the double digits but never resolves the consistent penny-scale offsets that keep it outside tolerance. Claude remains brittle across prompt levels, suggesting persistent misunderstandings of the dependency chain. As shown in Tables~\ref{tab:very_complex_levels_exact} and \ref{tab:very_complex_levels_mae}, prompt detail mitigates error magnitudes but perfection requires a model that can follow the full sequence without drift.

\begin{table}[H]
\centering
\small
\caption{Very complex dataset: Exact within tolerance (\%) by prompt level (L1 to L4). Values reflect the percent of rows within a 0.05 dollar tolerance after 2 decimal rounding.}
\label{tab:very_complex_levels_exact}
\begin{tabular}{lcccc}
\toprule
Model & L1 & L2 & L3 & L4 \\
\midrule
GPT~5~Auto & 0 & 0 & 8 & 29 \\
Claude~Sonnet~4 & 0 & 0 & 0 & 0 \\
Perplexity~Pro & 0 & 16 & 12 & 100 \\
Grok~Auto & 0 & 17 & 17 & 17 \\
\bottomrule
\end{tabular}
\end{table}

\begin{table}[H]
\centering
\small
\caption{Very complex dataset: Mean Absolute Error (USD) by prompt level (L1 to L4). Lower values indicate closer average magnitude to ground truth.}
\label{tab:very_complex_levels_mae}
\begin{tabular}{lcccc}
\toprule
Model & L1 & L2 & L3 & L4 \\
\midrule
GPT~5~Auto & 12044.12 & 1689.04 & 690.83 & 683.40 \\
Claude~Sonnet~4 & 974517.91 & 10219.30 & 10290.69 & 248.93 \\
Perplexity~Pro & 329.49 & 179.18 & 10555.43 & 0.00 \\
Grok~Auto & 399.76 & 28.93 & 28.93 & 28.93 \\
\bottomrule
\end{tabular}
\end{table}

Three insights emerge: (1) with linear arithmetic and short rule graphs, models perform nearly deterministically under minimal prompting; (2) for complex tasks, explicit formulas significantly improve accuracy, while partial narratives often reduce exactness; and (3) at very high complexity, cent-level tolerance reveals persistent rounding bias—only Perplexity achieves fully exact results, while GPT and Grok remain below 50\%, and Claude degrades likely due to context sensitivity. Gemini 2.5 Pro was tested but excluded due to recurring synthetic and schema issues.

\section{Discussion and Conclusion}

This study assessed whether LLMs can deliver payroll-grade performance in semantic understanding, ordered computation, and cent-level accuracy. Models showed strong semantic alignment across simple to complex datasets, reliably mapping fields and rules, but struggled in highly complex scenarios involving multi-jurisdiction rules, capped taxes, and FX interactions. Prompt-only approaches worked for simple dependencies, while tool use or programmatic computation significantly improved reliability for interacting rules by preserving order, thresholds, and auditability. Overall, LLMs are effective for straightforward payroll cases, but complex scenarios require explicit, reproducible computation. The framework identifies this break point and guides when prompting is sufficient versus when computation is necessary, with future work focused on stronger validation, sandboxed testing, and schema-drift safeguards.

\begin{acknowledgements}
This research was partially supported by the National Science Foundation under EPSCoR Award No. OIA-1946391.

\end{acknowledgements}

\bibliographystyle{spmpsci}  
\bibliography{bibliography.bib}

\end{document}